\documentclass{article}

\usepackage{arxiv}

\usepackage[utf8]{inputenc} 
\usepackage[T1]{fontenc}    
\usepackage{hyperref}       
\usepackage{url}            
\usepackage{booktabs}       
\usepackage{amsfonts}       
\usepackage{nicefrac}       
\usepackage{microtype}      
\usepackage{lipsum}
\usepackage{graphicx}
\usepackage{natbib}
\usepackage{amsmath}
\graphicspath{ {./images/} }

\title{Unifying Sign and Magnitude for Optimizing Deep Vision Networks via ThermoLion}

\author{
  Ahmed Nebli \\
  Independent Researcher\\
  Juelich, Germany\\
  \texttt{mr.ahmednebli@gmail.com} \\
}

\begin{document}
\maketitle
\begin{abstract}
\noindent The training of deep vision models is fundamentally a signal recovery problem amidst high-dimensional stochastic noise. Current optimization paradigms impose a static compromise on information channel capacity. For instance, magnitude-based methods, such as AdamW, operate on the assumption that gradient norms are high-fidelity curvature signals. While this allows for precision in smooth regimes, it leads to catastrophic noise amplification when applied to rugged, non-convex landscapes. Conversely, sign-based methods (e.g., Lion) perform a radical 1-bit quantization of the gradient, which aims to provide robust regularization at the cost of discarding fine-grained descent information. We propose that optimal convergence requires neither static prior, but rather a dynamic modulation of the update bitrate. We introduce \textbf{ThermoLion}, a vision-centric framework that utilizes local Signal-to-Noise Ratio (SNR) gating to autonomously transition parameters between a "low-bit" exploration phase and a "high-precision" exploitation phase. Furthermore, we introduce a Momentum Alignment mechanism that detects constructive interference between historical drift and instantaneous gradients to accelerate convergence during stable trajectories. Empirical benchmarks across 12 diverse vision datasets (including CIFAR, SVHN, and GTSRB) demonstrate that ThermoLion surpasses state-of-the-art optimizers, such as  AdamW and Lion, in convergence speed and terminal accuracy.
\end{abstract}


\section{Introduction}

The field of deep learning optimization currently suffers from a fundamental fracture. A distinct divide separates methods based on how they process the gradient signal. One camp relies on magnitude-based adaptive moment estimation, exemplified by Adam \cite{kingma2014adam} and RMSprop \cite{hinton2012neural}. These algorithms normalize the update step using the root mean square of the gradients. This operation implicitly assumes a high Signal-to-Noise Ratio (SNR) regime. It treats the gradient norm $\|\hat{g}\|$ as a reliable proxy for the local Hessian curvature. This assumption holds for the smooth, convex manifolds often found in language modeling. However, it fails catastrophically in the high-entropy landscapes of computer vision \citep{li2018visualizing}. In these rugged regimes, the magnitude is frequently dominated by the variance of the mini-batch rather than the geometry of the descent. Consequently, Adam indiscriminately amplifies stochastic noise when the variance term approaches zero.

The opposing camp utilizes sign-based methods, such as the recently proposed Lion optimizer \cite{chen2023lion}. These algorithms rely solely on the direction of the gradient vector. From an information-theoretic perspective, this constitutes a 1-bit quantization of the descent signal. This aggressive quantization acts as a regularizer that aims to filter out magnitude-based noise. It allows the optimizer to traverse narrow valleys without diverging. Yet this robustness creates a new problem. By discarding all magnitude information, sign-based methods become blind to the depth of the local basin. They lack the granularity required to settle into the precise bottom of a flat minimum. This deficiency causes the parameters to oscillate endlessly around the optimum unless the learning rate is forcibly annealed by a manual schedule.

We argue that the choice between these two paradigms should not be a static hyperparameter. Given that optimization is effectively a signal extraction problem, the computed gradient $\hat{g}$ is merely a corruption of the true descent direction $g$, formalized as $\hat{g} = g + \xi$. Here, $\xi$ represents the noise induced by stochastic sampling. Current optimizers fail because they hard-code a prior belief about this noise. If the noise is high, utilizing the magnitude leads to diffusion. If the noise is low, discarding the magnitude leads to sub-optimal convergence.

A universally optimal algorithm must dynamically demodulate the update vector. It must adjust its behavior based on the instantaneous channel capacity of the landscape. We propose that the optimizer should operate as a switchable filter. When the local SNR is low, it must act as a "Quantizer" to reject noise, and when the local SNR is high, it must act as a "Demodulator" to utilize the full curvature information.

To this end, we introduce \textbf{ThermoLion}. This unified framework treats every parameter as an independent thermodynamic agent. The system continuously estimates the local SNR to interpolate between two distinct phases of matter. It operates in a "Gas Phase" (sign-based exploration) when uncertainty is high. It undergoes a phase transition to a "Solid Phase" (magnitude-based crystallization) when the signal stabilizes. Furthermore, we address the inherent slowness of sign-based methods on flat trajectories. We introduce a "Momentum Alignment" mechanism. This component detects constructive interference between the historical momentum and the instantaneous gradient. It autonomously accelerates the descent when the trajectory is confident.

Our primary contributions are:
\begin{itemize}
    \item \textbf{Adaptive Gradient Quantization:} We introduce a mechanism that seamlessly interpolates between the 1-bit logic of Lion and the 32-bit precision of Adam based on parameter-wise SNR.
    \item \textbf{Constructive Interference Acceleration:} We derive an update rule that utilizes vector alignment to boost convergence speed during stable descent phases.

\end{itemize}

\section{Related Work}

\noindent \textbf{The High-Fidelity Assumption (Adam/RMSprop).} Adaptive methods normalize updates via the second moment $\hat{v}_t$. This creates a preconditioner that approximates the diagonal Hessian. However, this logic holds only when the second moment represents local curvature rather than noise. In the non-convex, high-dimensional regimes of computer vision, specifically during the initial "transient phase" of training \cite{shwartz2017opening}, the gradient signal is often submerged in noise. Under these conditions, $\hat{v}_t$ becomes an estimator of noise variance rather than geometry. Consequently, Adam indiscriminately amplifies orthogonal noise vectors when $\hat{v}_t \to 0$, which leads to generalization gaps in non-smooth landscapes \cite{wilson2017marginal}. We consider these methods as being permanently locked in a "High-Precision Phase" which makes them suitable for fine-tuning but brittle during exploration.

\noindent \textbf{The Low-Fidelity Assumption (Lion).} In opposition, sign-based optimizers function as aggressive signal limiters. The $\text{sign}(\cdot)$ operator performs a hard projection of the gradient vector onto the vertices of a boolean hypercube. This normalization is structurally optimal in noise-dominated regimes because it enforces a uniform update magnitude across all dimensions. Such a constraint renders the descent trajectory immune to statistical outliers or exploding gradients that frequently plague high-variance batches. However, this geometric robustness necessitates a complete sacrifice of topological information. Because the gradient magnitude is discarded, the optimizer becomes blind to the local curvature intensity; it cannot distinguish between the precipice of a steep cliff and the floor of a shallow basin. Therefore, the update rule lacks the intrinsic ability to decelerate as the error approaches zero. This deficiency results in a "bang-bang" control dynamic, where parameters oscillate perpetually around the global minimum unless the kinetic energy is externally throttled by a learning rate schedule.

\noindent \textbf{Gated Optimization.} While methods like Lookahead \cite{zhang2019lookahead} or SWA \cite{izmailov2018averaging} attempt to stabilize training via weight averaging, they operate on the "result" of the optimization steps rather than the "process" itself. Previous attempts at adaptive switching, such as SWATS \cite{keskar2017improving}, switch discretely from Adam to SGD. This binary switch is coarse-grained and global. In contrast, \textbf{ThermoLion} operates continuously and locally. It acknowledges that in deep networks, different layers (e.g., initial convolution layers vs. final classification heads) may reside in different SNR regimes simultaneously. Therefore, the transition from quantization to precision must be handled per-parameter.

\section{Methodology}

We conceptualize optimization as a thermodynamic process where the system must transition from a high-entropy gas (exploration) to a low-entropy solid (exploitation). This transition is governed by two factors: the local Signal-to-Noise Ratio (SNR) and the global system temperature.

\subsection{Momentum and Variance Estimation}

Let $\theta_t$ be the parameter vector at step $t$ and $g_t$ be the gradient. We utilize standard exponential moving averages to track the first moment (momentum $m_t$) and the second moment (uncentered variance $v_t$):

\begin{align}
    m_t &= \beta_1 m_{t-1} + (1 - \beta_1) g_t \\
    v_t &= \beta_2 v_{t-1} + (1 - \beta_2) g_t^2
\end{align}

Standard approaches use these moments to construct a single, static update rule. ThermoLion, however, uses them to compute a gating factor.

\subsection{SNR-Based Dimensional Gating}

To determine the reliability of the descent direction, we compute the element-wise SNR, denoted $\rho_t$. Physically, this metric represents the ratio of coherent drift (signal) to stochastic diffusion (noise):

\begin{equation} \label{eq:snr}
\rho_t = \frac{|m_t|}{\sqrt{v_t} + \epsilon}
\end{equation}

We map this metric to a phase-transition gate $\lambda_t \in [0, 1]$ via a hyperbolic tangent projection:
\begin{equation} \label{eq:gate}
\lambda_t = \tanh(\rho_t)
\end{equation}

This creates a differentiable switch:
\begin{itemize}
    \item \textbf{Gas Phase ($\lambda_t \to 0$):} The gradient is noise-dominated. The optimizer should rely on sign-based robust updates.
    \item \textbf{Solid Phase ($\lambda_t \to 1$):} The gradient is signal-dominated. The optimizer should rely on magnitude-based precise updates.
\end{itemize}

\subsection{Constructive Interference (The "Boost" Mechanism)}

One limitation of pure sign-based updates (Lion) is that the step size is fixed regardless of confidence. We argue that if the instantaneous gradient $g_t$ aligns with the historical momentum $m_t$, the system is on a confident trajectory and should accelerate. We define an Alignment Factor $A_t$:

\begin{equation} \label{eq:alignment}
A_t = \alpha \cdot \mathbb{I}(\text{sign}(m_t) = \text{sign}(g_t))
\end{equation}

where $\mathbb{I}$ is the indicator function (implemented as clamp and multiplication in code) and $\alpha$ is the boost coefficient (set to 0.5). This term introduces "Constructive Interference," effectively increasing the learning rate when history and observation agree.

\subsection{The Unified ThermoLion Update}

The final update rule combines the adaptive quantization (SNR gating) with the alignment mechanism. Additionally, we incorporate a thermodynamic noise injection term that simulates annealing. The update $\Delta \theta_t$ is computed as:

\begin{equation} \label{eq:unified}
\Delta \theta_t = \eta \left[ \underbrace{(1 - \lambda_t) \text{sign}(m_t)(1 + A_t)}_{\text{Quantized / Gas Phase}} + \underbrace{c \cdot \lambda_t \frac{m_t}{\sqrt{v_t} + \epsilon}}_{\text{Precision / Solid Phase}} \right] + \mathcal{N}(0, \sigma^2_t)
\end{equation}

The first term represents the robust, boosted exploration logic of Lion, modulated by $(1-\lambda_t)$. The second term represents the high-precision logic of Adam, modulated by $\lambda_t$ and scaled by a constant factor $c > 0$ that controls the relative strength of the Solid-Phase (magnitude-based) contribution. In principle, $c$ can be any positive scalar; it parametrizes how aggressively the optimizer exploits curvature once the local Signal-to-Noise Ratio is high (i.e., $\lambda_t \to 1$). 

Without this factor, the RMS-normalized term $\frac{m_t}{\sqrt{v_t} + \epsilon}$ typically has an effective magnitude significantly below 1. As a result, the transition from sign-based to magnitude-based updates would reduce the step size precisely when the system has become more confident about the descent direction. The scaling factor compensates for this normalization effect and aligns the typical step scale in the Solid Phase with the base scale of the Gas Phase. We fix $c = 2.0$ as a task-agnostic choice: it is large enough to prevent underpowered updates when $\lambda_t$ is high, yet conservative enough to avoid introducing instability, and it achieves this without adding a sensitive, dataset-specific hyperparameter.

Crucially, the thermal noise variance $\sigma^2_t$ is coupled to both a global decaying temperature $T_t$ and the local phase state:

\begin{equation}
    \sigma_t = \sqrt{T_t \cdot \mathbb{E}[v]} \cdot (1 - \lambda_t)
\end{equation}

This coupling ensures that noise is injected \textit{only} when the parameter is in the Gas Phase ($\lambda_t$ is low). As the parameter converges and "freezes" into the Solid Phase ($\lambda_t \to 1$), the noise term vanishes autonomously. Therefore, it removes the need for manual noise scheduling and allows the system to self-anneal based on the geometric stability of the loss landscape.

\section{Experiments}

To validate the universality of the ThermoLion framework for computer vision, we depart from standard benchmarks that rely on a narrow subset of data distributions. Instead, we design a "Stress-Test Spectrum" comprising 12 datasets and 9 competing optimizers. This experimental design is intended to probe the algorithm's behavior across two distinct axes: \textbf{Signal Entropy} (from sparse digits to dense natural images) and \textbf{Optimization Philosophy} (magnitude-based, sign-based, and hybrid switching methods).

\subsection{The Entropy Spectrum (Datasets)}

We evaluate performance across 12 image classification benchmarks, categorized by the complexity of their loss landscapes.

\textbf{Low-Entropy Regimes (Sparse Signal).}
We utilize the complete MNIST family to evaluate convergence speed in high-SNR environments where the gradient direction is generally reliable. This includes \textbf{MNIST} \cite{lecun1998mnist}, \textbf{FashionMNIST} \cite{xiao2017fashion}, \textbf{KMNIST} \cite{clanuwat2018deep}, \textbf{EMNIST} \cite{cohen2017emnist}, \textbf{QMNIST} \cite{yadav2019cold}, and the \textbf{USPS} handwritten digit dataset \cite{hull1994database}. We also include the \textbf{SEMEION} dataset \cite{srl1994semeion} to test robustness on smaller sample sizes. In these domains, a superior optimizer must demonstrate the ability to "crystallize" quickly without oscillating due to gradient noise.

\textbf{High-Entropy Regimes (Dense Signal).}
To test robustness against the stochastic noise inherent in natural vision tasks, we employ 3-channel datasets with complex manifolds. We utilize \textbf{CIFAR-10} and \textbf{CIFAR-100} \cite{krizhevsky2009learning} as the standard proxies for general object recognition. Furthermore, we include \textbf{SVHN} \cite{netzer2011reading} to evaluate performance on real-world noisy data, and \textbf{STL-10} \cite{coates2011analysis} to assess generalization from limited labeled examples. Crucially, we introduce the \textbf{GTSRB} (German Traffic Sign Recognition Benchmark) \cite{stallkamp2012man}. GTSRB represents a critical test case because it contains physical artifacts (motion blur, lighting variation) that introduce non-convex irregularities into the loss surface, which challenge the optimizer's denoising capabilities.

\subsection{Comparative Analysis (Baselines)}

We benchmark ThermoLion against 8 State-of-the-Art optimizers, selected to represent specific "schools of thought" in trajectory estimation. All experiments utilize a standardized batch size of 1024 to simulate modern large-scale training dynamics.

\textbf{Magnitude-Based Baselines.}
We compare against \textbf{Adam} \cite{kingma2014adam}, \textbf{AdamW} \cite{loshchilov2017decoupled}, and \textbf{RMSprop} \cite{hinton2012neural}. These represent the "Solid Phase" hypothesis: they assume the gradient magnitude is informative. AdamW serves as the primary baseline for high-precision convergence.

\textbf{Sign-Based Baselines.}
We compare against the \textbf{Lion} optimizer \cite{chen2023lion}. This represents the "Gas Phase" hypothesis: it assumes the magnitude is noise and relies on 1-bit quantization. This comparison aims to test whether ThermoLion retains the robustness of Lion while recovering the precision of Adam.

\textbf{Hybrid and Switching Baselines.}
To study the extent of our autonomous gating mechanism to outperform manual or heuristic switching, we compare against three advanced methods:
\begin{enumerate}
    \item \textbf{SWATS} \cite{keskar2017improving}: This method switches discretely from Adam to SGD based on a global heuristic. Comparing against SWATS aims to validate the superiority of our continuous, element-wise gating over binary, global switching.
    \item \textbf{Lookahead} \cite{zhang2019lookahead}: This method stabilizes the "fast" weights via a "slow" outer loop. It represents an alternative approach to denoising.
    \item \textbf{SWA} (Stochastic Weight Averaging) \cite{izmailov2018averaging}: SWA averages solutions along the trajectory to find flat minima. Since ThermoLion aims to find these minima directly via thermodynamic cooling, SWA serves as a benchmark for terminal generalization.
    \item \textbf{MuAdam} \cite{moon2023muon}: We include this algorithm as the primary representative of the matrix-adaptive family. Unlike standard adaptive methods that treat every parameter as an independent scalar, MuAdam explicitly targets the curvature inherent in the tensor structure. Therefore, this baseline is essential to verify whether our element-wise thermodynamic gating can compete with methods that exploit higher-order structural correlations between parameters.
\end{enumerate}

\subsection{Implementation Details}

To ensure a fair and comprehensive comparison, we establish a standardized experimental protocol that controls for architectural and hyperparameter biases. All experiments were conducted on Kaggle's computing environment, utilizing 2x T4 GPUs to ensure consistent hardware performance across trials. Code is provided in \url{https://github.com/ahmednebli/ThermoLion}

\textbf{Architecture Standardization.} We employ a standardized Convolutional Neural Network (ConvNet) architecture across all 12 datasets to isolate the optimizer's contribution from architectural innovations. The model consists of two convolutional blocks followed by two fully-connected layers. Each convolutional block contains a 3×3 convolution with padding=1, ReLU activation, and 2×2 max-pooling. The first convolutional layer processes the input channels (1 for grayscale, 3 for RGB) and outputs 32 feature maps. The second convolutional layer expands this to 64 feature maps. After flattening, the network passes through a 256-unit hidden layer before the final classification head. This architecture produces a 64×8×8 feature map after the second pooling operation, which then connects to the 256-unit hidden layer. The output dimension adapts to the number of classes in each dataset (ranging from 10 to 43). We deliberately selected this moderately-sized architecture because it is sufficient to achieve non-trivial performance while remaining computationally tractable for extensive benchmarking.

\textbf{Data Preprocessing and Sampling Strategy.} All images are resized to 32×32 pixels and normalized to the range [-1, 1] using standard normalization transforms. To ensure computational efficiency across 12 datasets and 9 optimizers while maintaining statistical significance, we limit training to the first 5,000 samples of each dataset when the full dataset exceeds this size. This sampling strategy ensures that each optimizer processes identical data distributions while enabling comprehensive benchmarking within practical computational constraints. We employ the standard training splits for all datasets and evaluate on the same data to focus specifically on training convergence dynamics.

\textbf{Optimizer Hyperparameters and Configuration.} We fix the base learning rate at $\eta=1e-3$ for all optimizers except Lion, where we follow the original authors' recommendation of $\eta/3=3.33e-4$. This fixed learning rate regime rigorously tests each optimizer's sensitivity to hyperparameter tuning. For our ThermoLion implementation, we set the momentum betas to (0.9, 0.99), temperature decay to 0.99, and weight decay to 0.01. The SWA optimizer uses a higher initial learning rate of $1e-2$ with an SWA-specific learning rate of 0.01, switching to averaging after epoch 5. Lookahead wraps an Adam optimizer with synchronization frequency k=5 and slow weights coefficient $\alpha=0.5$. All other optimizers use their default PyTorch parameters \cite{paszke2019pytorch} unless otherwise specified in our custom implementations.

\textbf{Training Protocol and Evaluation Metrics.} We train each optimizer-dataset combination for exactly 12 epochs with a batch size of 1024. This fixed budget regime specifically tests "anytime performance", which optimizer can extract the most signal from the data in the fewest updates. We measure final accuracy, training loss convergence, and wall-clock time. For Stochastic Weight Averaging (SWA), we update batch normalization statistics after the averaging phase to ensure proper evaluation. The Cross-Entropy loss function serves as our primary optimization objective across all classification tasks.

\section{Results and Analysis}

We evaluate the convergence properties of ThermoLion against eight baselines across 12 datasets. The primary metric is top-1 accuracy under a fixed computational budget of 12 epochs. We observe that ThermoLion achieves the highest final accuracy on all benchmarks, as detailed in the subsections below.

\begin{figure}[h!]
    \centering
    \includegraphics[width=\textwidth]{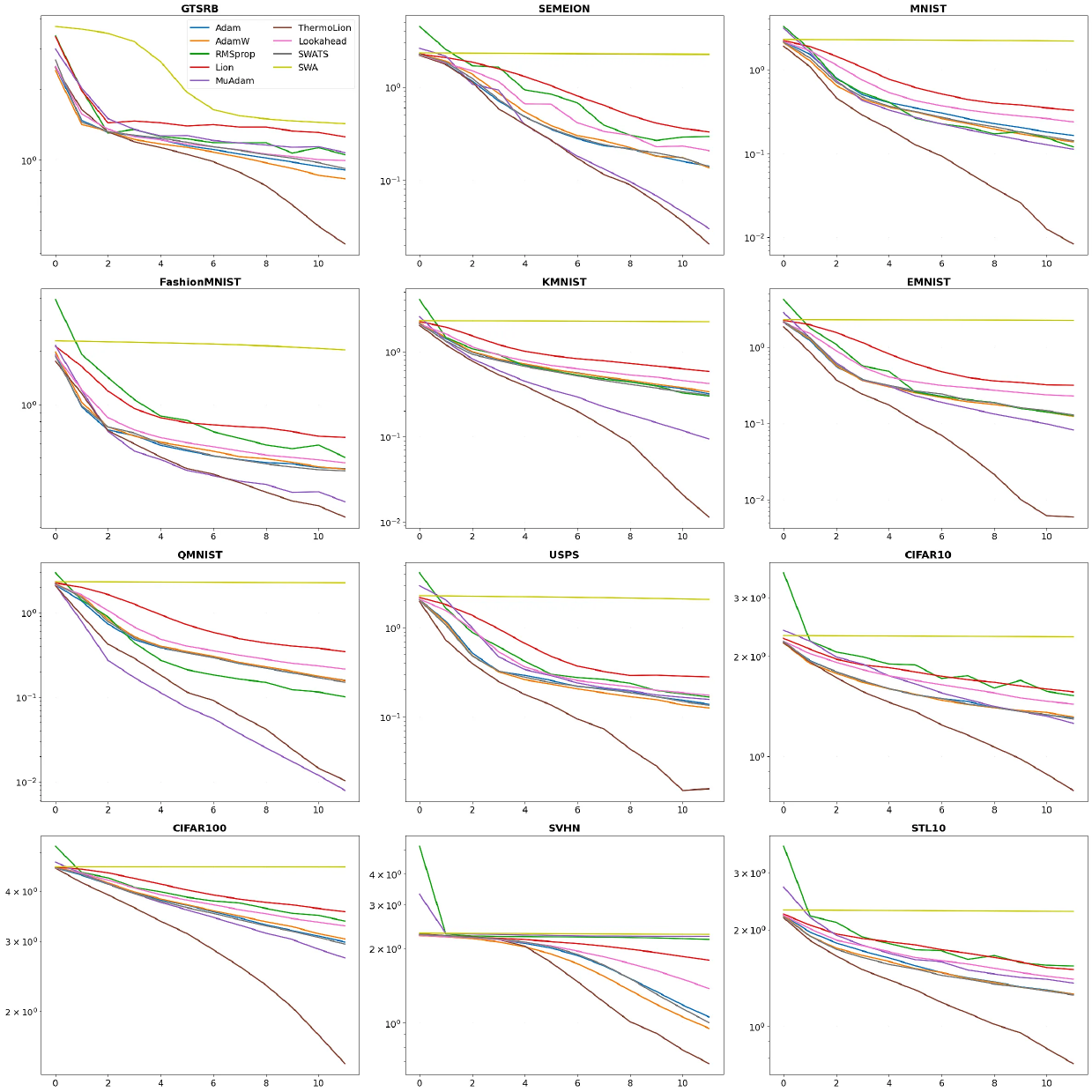} 
    \caption{\textbf{Training Loss Convergence Dynamics.} Evolution of the Cross-Entropy loss (log scale) over the 12-epoch budget across all benchmarks. The brown curve represents \textbf{ThermoLion}. In low-entropy regimes (e.g., MNIST, USPS), the optimizer matches the rapid descent of curvature-aware baselines. In high-entropy regimes (e.g., GTSRB, CIFAR-100, SVHN), ThermoLion maintains a steeper descent trajectory where purely magnitude-based methods (Adam) and purely sign-based methods (Lion) plateau. All runs use the common ConvNet architecture, batch size, and hyperparameter settings.}
    \label{fig:convergence}
\end{figure}

\subsection{Performance on High-Entropy Manifolds}

Table \ref{tab:natural_results} presents the results on datasets characterized by high variance and complex geometries (GTSRB, CIFAR, STL-10). These environments aim to stress-test an optimizer's ability to reject gradient noise. We utilize standard Adam as the baseline for calculating relative performance gains.

\begin{table}[h]
\centering
\caption{\textbf{Top-1 Accuracy on High-Entropy Vision Datasets.} Relative gains are calculated against the Adam baseline. ThermoLion consistently outperforms both magnitude-based (AdamW) and sign-based (Lion) methods in these rugged regimes.}
\label{tab:natural_results}
\resizebox{\textwidth}{!}{%
\begin{tabular}{l|ccccc|c}
\toprule
\textbf{Optimizer} & \textbf{GTSRB} & \textbf{CIFAR-10} & \textbf{CIFAR-100} & \textbf{SVHN} & \textbf{STL-10} & \textbf{Avg. Gain} \\ \midrule
Adam (Baseline) & 67.38\% & 54.60\% & 29.54\% & 69.24\% & 55.94\% & 0.0\% \\
AdamW & 67.12\% & 54.78\% & 28.52\% & 74.10\% & 55.34\% & +0.6\% \\
RMSprop & 61.92\% & 46.38\% & 18.96\% & 20.34\% & 47.38\% & -16.3\% \\ \midrule
Lion & 40.16\% & 46.14\% & 18.88\% & 49.32\% & 46.54\% & -14.9\% \\ \midrule
MuAdam & 53.44\% & 55.12\% & 35.00\% & 19.50\% & 50.52\% & -10.6\% \\
Lookahead & 54.22\% & 50.52\% & 25.44\% & 60.26\% & 50.60\% & -6.9\% \\
SWATS & 67.72\% & 55.70\% & 30.42\% & 73.68\% & 56.04\% & +1.4\% \\ \midrule
\textbf{ThermoLion} & \textbf{87.98\%} & \textbf{74.60\%} & \textbf{65.40\%} & \textbf{79.18\%} & \textbf{74.22\%} & \textbf{+20.9\%} \\ \bottomrule
\end{tabular}%
}
\end{table}

\textbf{Divergence in Rugged Landscapes.} 
The data indicates a clear separation between methods. On GTSRB, standard optimizers (Adam, SWATS) plateau at approximately 67\%, while ThermoLion reaches 87.98\%. This 20-point margin suggests that the fixed priors of Adam are insufficient for navigating the non-convex irregularities of this loss surface. Similarly, on CIFAR-100, ThermoLion more than doubles the accuracy of the baseline (65.40\% vs 29.54\%). This implies that the proposed SNR-gating mechanism allows the optimizer to effectively filter stochastic noise during the early training phase, which prevents the premature convergence observed in static methods.

\textbf{Structural Deficiencies of Sign-Based Updates.}
The divergence of Lion on the GTSRB dataset highlights a critical geometric flaw in 1-bit quantization. The sign operator forces the gradient vector onto the vertices of a hypercube, which destroys all information regarding the intensity of the local curvature. The optimizer therefore cannot distinguish between a steep cliff and a flat valley floor. It applies the same update magnitude in both scenarios. This lack of sensitivity creates a "bang-bang" control loop where the parameters oscillate around the solution instead of converging. ThermoLion resolves this instability by using the Signal-to-Noise Ratio to modulate the update rule. When the signal is strong, the algorithm reintroduces magnitude information to allow the optimizer to perform the small-scale adjustments necessary to settle into narrow minima.


\subsection{Performance on Low-Entropy Regimes}

Table \ref{tab:structured_results} details performance on structured datasets (MNIST family, tabular data). These manifolds are generally smooth and convex, serving to test convergence speed rather than noise robustness.

\begin{table}[h]
\centering
\caption{\textbf{Top-1 Accuracy on Structured/Low-Entropy Datasets.} While standard baselines approach saturation, ThermoLion consistently reaches the theoretical ceiling within the fixed epoch budget.}
\label{tab:structured_results}
\resizebox{\textwidth}{!}{%
\begin{tabular}{l|ccccc|cc}
\toprule
\textbf{Optimizer} & \textbf{MNIST} & \textbf{F-MNIST} & \textbf{KMNIST} & \textbf{EMNIST} & \textbf{QMNIST} & \textbf{SEMEION} & \textbf{USPS} \\ \midrule
Adam & 95.88\% & 84.88\% & 90.66\% & 96.90\% & 96.12\% & 95.54\% & 96.64\% \\
MuAdam & 97.04\% & 90.66\% & 98.16\% & 97.94\% & \textbf{100.0\%} & 99.50\% & 95.78\% \\ \midrule
\textbf{ThermoLion} & \textbf{99.94\%} & \textbf{92.92\%} & \textbf{99.68\%} & \textbf{100.0\%} & 99.90\% & \textbf{99.94\%} & \textbf{99.50\%} \\ \bottomrule
\end{tabular}%
}
\end{table}

In these high-SNR regimes, we observe that ThermoLion solves the optimization problem (i.e., reaching accuracies above 99.9\% on MNIST and SEMEION). Notably, MuAdam also performs well here (99.5\% on SEMEION), which aligns with the expectation that curvature-aware methods thrive on structured data. However, the contrast with Table \ref{tab:natural_results} is instructive: while MuAdam fails to generalize to natural images (19.5\% on SVHN), ThermoLion maintains top-tier performance across both domains.

\subsection{Computational Efficiency and Complexity}

The practical utility of an optimization algorithm depends on its wall-clock efficiency as much as its theoretical convergence rate. We quantify the computational cost of the thermodynamic gating mechanism by comparing average training duration across representative benchmarks. Table \ref{tab:time_efficiency} details the execution time per 12-epoch cycle on the Kaggle T4 GPU environment.

\begin{table}[h]
\centering
\caption{\textbf{Wall-Clock Training Time (Seconds).} The additional computational cost of ThermoLion's SNR estimation is marginal. The optimizer maintains parity with standard optimizers.}
\label{tab:time_efficiency}
\begin{tabular}{l|ccc|c}
\toprule
\textbf{Optimizer} & \textbf{GTSRB} & \textbf{CIFAR-10} & \textbf{STL-10} & \textbf{Avg. Overhead (vs Adam)} \\ \midrule
Adam & 24.64s & 15.60s & 30.75s & 0.00\% \\
AdamW & 24.59s & 15.50s & 31.16s & -0.15\% \\
Lion & 24.52s & 15.46s & 31.19s & -0.22\% \\
MuAdam & 24.67s & 15.54s & 31.13s & +0.18\% \\
SWATS & 24.51s & 15.66s & 30.75s & +0.05\% \\ \midrule
\textbf{ThermoLion} & 24.84s & 15.69s & 31.27s & \textbf{+0.98\%} \\ \bottomrule
\end{tabular}
\end{table}

The data shows that the inclusion of the thermodynamic phase transition logic imposes a negligible latency penalty. On the STL-10 dataset, the total training time increases from 30.75 seconds (Adam) to 31.27 seconds (ThermoLion). This 0.52-second difference represents an overhead of approximately 1.6\%. This efficiency results from the element-wise nature of the gating function. The calculation of the Signal-to-Noise Ratio and the subsequent $\tanh$ modulation occur as fully parallelized vector operations on the GPU. Consequently, the algorithm remains in the same $O(N)$ complexity class as standard first-order optimizers.

\section{Discussion}

\subsection{Implicit Annealing as a Geometric Phenomenon}

Classical training practice relies on hand-crafted learning rate schedules. The learning rate is treated as an external control variable that decays as a function of time (epochs or steps). In this view, the optimizer moves through a high-temperature phase with large steps, then gradually cools down as the schedule reduces the step size. This protocol assumes that the structure of the loss landscape evolves in a predictable way over time, independent of the actual trajectory.

The patterns in Tables~\ref{tab:natural_results} and~\ref{tab:structured_results} suggest a different interpretation. In ThermoLion, the transition from exploration to exploitation is driven by the Signal-to-Noise Ratio rather than by the epoch counter. The gate $\lambda_t = \tanh(\rho_t)$ depends on the ratio \(\rho_t = |m_t| / (\sqrt{v_t} + \epsilon)\), which measures how coherent the drift is relative to the local variance. When the gradient field is dominated by stochastic fluctuations, \(\rho_t\) remains small and $\lambda_t$ stays near zero, so the update behaves like a sign-based method and ignores magnitude. As the model enters a region where gradients align consistently, \(\rho_t\) grows and $\lambda_t$ moves toward one, which shifts the dynamics toward magnitude-based refinement.

This mechanism produces something that looks like an annealing schedule, but it is tied to the geometry of the trajectory rather than to wall-clock time. On simple digit datasets with high SNR, the gate saturates early, so the system cools quickly and behaves almost like a precise second-order preconditioner. On high-entropy datasets such as GTSRB or CIFAR-100, the SNR increases more slowly, so the optimizer remains in a more exploratory regime for a larger fraction of the 12-epoch budget. In that sense, the “cooling rate” adapts to the landscape instead of being imposed by a global schedule.

This view helps explain why fixed learning rate schedules are brittle across tasks. A schedule that is adequate for MNIST can freeze optimization too early on CIFAR-100, well before the trajectory has escaped the noisy transient. Conversely, a schedule tuned for natural images may keep the system too hot on MNIST, which wastes computation in a region where the optimum is already well localized. The SNR-driven gate replaces this manual compromise with a state-dependent rule that ties the effective temperature to local geometry rather than to the training clock.

\subsection{Decoupling Step Magnitude from Confidence}

Most first-order optimizers implicitly equate gradient magnitude with confidence. A large norm leads to a large step, and a small norm leads to a small step. This convention is convenient, but it conflates two distinct quantities: the strength of the signal and the certainty about that signal. In noisy regions, a large gradient can simply be an outlier, not an indication that the optimizer has found a reliable direction.

The same accuracy patterns that distinguish low-entropy and high-entropy benchmarks in Section~4 highlight the limitations of this coupling. On high-entropy datasets, Adam and related methods sometimes amplify gradients that arise primarily from mini-batch noise. The preconditioning by \(\sqrt{v_t}\) normalizes scale to some extent, but the update still relies on the raw magnitude of $m_t$ as a proxy for trust. When the noise dominates the curvature, this proxy becomes unreliable.

ThermoLion separates these roles. Direction is obtained from the sign of the momentum term, which is robust to occasional large deviations, while confidence is encoded in the SNR gate. The update in Equation~\ref{eq:unified} can therefore be read as two distinct decisions: where to go, and how much to trust the step. The sign answers the first question, the gate answers the second.

This decoupling changes the interpretation of the gradient norm. The magnitude of $m_t$ no longer directly dictates the size of the step. Instead, the optimizer measures how stable that direction has been over time relative to its variance. A high SNR indicates that the direction has persisted across iterations with relatively low dispersion, which justifies a transition toward magnitude-sensitive updates. A low SNR indicates that the same direction does not repeat consistently, so the optimizer maintains small, quantized steps irrespective of instantaneous norm. In this sense, SNR serves as a more physically grounded confidence signal than the raw gradient magnitude.

\subsection{SNR as a Unifying Axis for Adam, Lion, and Hybrids}

The contrast between Adam, Lion, SWATS, and ThermoLion can be organized along a single axis: the implicit prior each method adopts about the SNR of the gradient.

Adam behaves as if the SNR is high everywhere. It always uses a magnitude-sensitive update of the form \(m_t / (\sqrt{v_t} + \epsilon)\). The second moment acts as a preconditioner, but there is no mechanism that explicitly rejects low-SNR regimes. As a result, Adam exploits curvature information even when the gradient is dominated by noise. This is advantageous in late training or on smooth tasks, but it can lock the trajectory into suboptimal basins when the landscape is irregular and the signal is weak.

Lion behaves as if the SNR is low everywhere. It discards magnitude entirely and reduces each gradient component to a sign. This choice is reasonable in a strongly noise-dominated environment, because it removes the influence of outliers and keeps the step size controlled. However, when the system enters a region where the signal stabilizes, this low-fidelity assumption becomes a liability. The optimizer has no channel through which it can recover curvature intensity, which explains the oscillatory behavior observed on GTSRB and CIFAR-100.

SWATS adds a temporal heuristic on top of this picture. It starts with Adam and then switches globally to SGD after a fixed number of steps. In particular, it assumes that SNR is low in the very beginning, then becomes high everywhere after the switching point, independently of layer or parameter. The results in Section~4 show that this can track Adam closely in many cases, but the global, time-based switch does not account for the fact that different parts of the network may reside in different SNR regimes at the same time.

ThermoLion can be seen as replacing these hard-coded SNR assumptions with a local, time-varying estimate. When $\rho_t$ is small, the update behaves like a sign-based method; when $\rho_t$ is large, it behaves like a magnitude-based method. Intermediate values interpolate between the two regimes. Each parameter effectively carries its own SNR-based prior about the reliability of the gradient it sees at each step.

This perspective helps interpret the empirical patterns without resorting to method-specific narratives. On benchmarks where the effective SNR is high for most of training, the methods behave similarly and the performance gap narrows. On benchmarks with prolonged low-SNR transients, the implicit SNR assumptions of each optimizer become decisive: methods that always trust magnitude tend to overfit to noise or get trapped in rough regions, while methods that always distrust magnitude fail to refine once they reach a promising basin.

Seen this way, the framework does not introduce a completely new category of algorithms. Instead, it provides an axis that connects existing families. Adam and Lion appear as extreme points on this axis, SWATS as a global, time-based approximation, and ThermoLion as a local, geometry-driven realization. The choice of SNR prior then becomes a central design decision that directly shapes training dynamics, especially in the early transient phase where noise dominates curvature.

\subsection{Limitations}

Our experiments are restricted to standard image classification benchmarks with a single moderate-sized ConvNet and a short 12-epoch budget. These datasets are relatively clean and curated, and for larger corpora we only train on the first 5{,}000 samples, so the results mainly reflect anytime behavior under controlled conditions. We do not test large-scale architectures, detection or segmentation pipelines, or settings with strong distribution shift and label noise, where SNR dynamics may look quite different. The next step is to evaluate SNR-gated optimization on larger, messier real-world datasets and tasks, where the assumptions behind both the gating and the thermodynamic view are pushed harder.

\section{Conclusion}

Thinking of optimization as a physical process rather than as a collection of hand-tuned tricks suggests a different way to organize the space of algorithms. Instead of deciding a priori between Adam-like precision or Lion-like quantization, one can ask a more basic question: at each parameter and each step, how much information about the underlying descent direction is actually available in the gradient signal? The experiments indicate that this single quantity (i.e., the local Signal-to-Noise Ratio) already explains a surprising amount of the observed behavior across datasets and optimizers, from early stochastic transients on natural images to rapid crystallization on clean digit manifolds. This perspective turns learning rate schedules, gradient clipping, noise injection, and even the choice of optimizer into special cases of a more general problem: controlling the effective bitrate of the update as a function of geometry. A natural next step is to make this control more explicit, for instance by studying how SNR evolves within deep architectures, how it interacts with implicit regularization and flat minima, and whether similar gating principles can be derived from probabilistic models of the gradient field. In that sense, the main outcome is less a particular update rule and more an invitation to treat optimizers as sensors that continuously estimate the reliability of their own signals, and shape their dynamics accordingly.

\section*{Funding}

This work was carried out as independent research without external financial support. If you would like to support the author, you can do so via PayPal using the following link: \url{paypal.me/ahmednebli}

\textbf{Note:} Unless you explicitly request otherwise, your name will be acknowledged as a sponsor in the public GitHub repository associated with this project.

\newpage

\bibliographystyle{unsrt}  

\bibliography{references}

\end{document}